\documentclass[10pt,twocolumn,letterpaper]{article}

\usepackage{cvpr}              

\definecolor{cvprblue}{rgb}{0.21,0.49,0.74}
\usepackage[pagebackref,breaklinks,colorlinks,allcolors=cvprblue]{hyperref}
\usepackage{url}            
\usepackage{booktabs}       
\usepackage{amsfonts}       
\usepackage{nicefrac}       
\usepackage{microtype}      
\usepackage{xcolor}         
\usepackage{colortbl}
\usepackage{pifont}
\usepackage{listings}
\newcommand{\cmark}{\ding{51}}
\newcommand{\xmark}{\ding{55}}
\usepackage{multirow}
\usepackage{bm}
\usepackage{amsthm}
\usepackage{graphicx}
\usepackage{amsmath}
\usepackage{subcaption}
\usepackage{enumitem}
\usepackage{amssymb}
\usepackage{mathtools}
\usepackage{overpic}
\usepackage{xspace}
\usepackage{wrapfig}
\newcommand{\CC}[1]{\cellcolor{gray!#1}}
\usepackage{pifont}
\usepackage{tocloft}
\usepackage[toc,page,header]{appendix}
\usepackage{adjustbox}
\usepackage{minitoc}

\newcommand{\alg}{\textrm{H-CLIP}\xspace}
\newcommand{\blg}{\textrm{POP}\xspace}
\usepackage{xcolor}         
\usepackage{soul}
\usepackage{marvosym}
\definecolor{frenchblue}{rgb}{0.0, 0.45, 0.73}

\def\paperID{****} 
\def\confName{CVPR}
\def\confYear{2025}

\title{Parameter-efficient Fine-tuning in Hyperspherical Space for Open-vocabulary Semantic Segmentation}

\begin{document}
\author{Zelin Peng$^{1}$~ Zhengqin Xu$^{1}$~ Zhilin Zeng$^{1}$~ Yaoming Wang$^{2}$~ Wei Shen$^{1{(\textrm{\Letter})}}$\\
	$^1$Shanghai Jiao Tong University\\  $^2$Huawei Inc.\\
	{\tt\small \{zelin.peng, fate311, bernardeschi, wang\_yaoming, wei.shen\}@sjtu.edu.cn;} 
}

\maketitle

\newcommand\blfootnote[1]{%
\begingroup 
\renewcommand\thefootnote{}\footnote{#1}%
\addtocounter{footnote}{-1}%
\endgroup 
}
\blfootnote{$^{\textrm{\Letter}}$ Corresponding Author: \texttt{wei.shen@sjtu.edu.cn}}

\begin{abstract}
Open-vocabulary semantic segmentation seeks to label each pixel in an image with arbitrary text descriptions. Vision-language foundation models, especially CLIP, have recently emerged as powerful tools for acquiring open-vocabulary capabilities. 
However, fine-tuning CLIP to equip it with pixel-level prediction ability often suffers three issues: 1) high computational cost, 2) misalignment between the two inherent modalities of CLIP, and 3) degraded generalization ability on unseen categories. To address these issues, we propose \alg, a symmetrical parameter-efficient fine-tuning (PEFT) strategy conducted in hyperspherical space for both of the two CLIP modalities. Specifically, the PEFT strategy is achieved by a series of efficient block-diagonal learnable transformation matrices and a dual cross-relation communication module among all learnable matrices. Since the PEFT strategy is conducted symmetrically to the two CLIP modalities, the misalignment between them is mitigated. Furthermore, we apply an additional constraint to PEFT on the CLIP text encoder according to the hyperspherical energy principle, i.e., minimizing hyperspherical energy during fine-tuning preserves the intrinsic structure of the original parameter space, to prevent the destruction of the generalization ability offered by the CLIP text encoder. Extensive evaluations across various benchmarks show that H-CLIP achieves new SOTA open-vocabulary semantic segmentation results while only requiring updating approximately 4\% of the total parameters of CLIP.
\end{abstract}
\section{Introduction}
\label{sec.1}

The aim of open-vocabulary semantic segmentation is to create a segmentation model capable of labeling each pixel in an image with categories that are not limited to a specific closed set according to text descriptions. Vision-language foundation models~\cite{VLM_lxmert_arxiv_2019,VLM_uniter_ECCV_2020,VLM_vilbert_NIPS_2019,CLIP_PMLR_2021,VLM_end_cvpr_2022,VLM_see_2021_CVPR,VLM_BLIP_2022_ICML,VLM_vilt_2021_ICML,VLM_align_nips_2021,VLM_large_2020_nips,VLM_mdetr_2021_ICCV,VLM_oscar_ECCV_2020,VLM_coarse_nips_2022,VLM_ground_cvpr_2022,VLM_UFO_2021_arxiv}, especially CLIP~\cite{CLIP_PMLR_2021}, are often utilized to endow open-vocabulary recognition capabilities. Consequently, open-vocabulary semantic segmentation essentially boil down to transferring these vision-language foundation models, originally trained with image-level supervision, to perform pixel-level predictions.

To this end, current methods~\cite{fcclip_nips_2023,SED_cvpr_2024,catseg_cvpr_2024,SAN_CVPR_2023} typically fine-tune CLIP on a benchmark dataset with segmentation annotations, i.e., COCO~\cite{COCO_2018_CVPR}, to equip it with the segmentation ability. However, this often leads to three main issues. First, fine-tuning CLIP on limited categories would affect its generalization ability, resulting in significant performance degradation on unseen categories. Second, current fine-tuning methods for open-vocabulary segmentation are usually asymmetrical~\cite{Lseg_ICLR_2022,SED_cvpr_2024,SAN_CVPR_2023}, i.e., typically freezing CLIP's text encoder and fine-tuning its image encoder. This strategy inevitably causes a potential obstacle: misalignment. More specifically, the misalignment arises from different alignment granularities. The text encoder maintains image-to-text alignment, while the image encoder shifts from image-to-text to pixel-to-text alignment. Due to these different alignment goals, the optimization process is largely impeded, leading to sub-optimal performance. Third, although remarkable performance gains, these approaches often rely on computationally extensive full fine-tuning, which raises concerns about scalability and affordability.

To address these issues, we propose a symmetric parameter-efficient fine-tuning strategy for CLIP, dubbed H-CLIP. First, to preserve CLIP's generalization ability, we are inspired by the hyperspherical energy principle~\cite{HE_nips_2018,OFT_nips_2023}, which suggests that maintaining the same hyperspherical energy during fine-tuning preserves the intrinsic structure, i.e., generalization capability. Therefore, H-CLIP is fine-tuned in hyperspherical space, incorporating orthogonal constraints in the learnable matrices of CLIP's text encoder. These orthogonal transformations are realized in a block-diagonal form to efficiently maintain the invariance of hyperspherical energy during fine-tuning. Second, by fine-tuning both modalities of CLIP with a comparable number of parameters, we largely mitigate the misalignment. Then, to further enhance alignment between different modalities of CLIP, we introduce a dual cross-relation communication (DCRC) module to explicitly encourage cross-modal and cross-layer communications within all learnable matrices. Notably, DCRC encourages the alignment problem with minimal overhead. Consequently, H-CLIP can adapt CLIP for open vocabulary semantic segmentation with only a small number of introduced learnable parameters.

Extensive results demonstrate that \alg achieves new state-of-the-art open-vocabulary semantic segmentation results across three benchmarks by fine-tuning CLIP with approximately 4\% of the total parameters of CLIP.

\section{Related Work}

\subsection{Open-vocabulary Semantic Segmentation}
Prior open-vocabulary semantic segmentation works typically perform this task through leveraging CLIP~\cite{CLIP_PMLR_2021}. initial efforts like~\cite{Freeclip_2022_ECCV} directly fine-tune CLIP on mainstream segmentation datasets, e.g., COCO~\cite{COCO_2018_CVPR}. However, they claim that fine-tuning CLIP's encoder significantly reduces its ability to generalize to unseen classes. To address this issue, some methods~\cite{Openseg_ECCV_2022,ZegFormer_CVPR_2022,ZSseg_ECCV_2022,ODISE_CVPR_2023} swing to the opposite extreme, fine-tuning an additional mask generator~\cite{mask2former_2022_CVPR} for segmentation while keeping CLIP frozen to maintain generalization-oriented recognition. However, this frozen parameter space lacks segmentation awareness, resulting in a misalignment between regions and text descriptions~\cite{OVSeg_CVPR_2023}. Other studies~\cite{fcclip_nips_2023,SAN_CVPR_2023,catseg_cvpr_2024} propose an advanced solution that fine-tunes only selected parameters, e.g., certain layers of CLIP, to enable pixel-level predictions while keeping most of CLIP's parameters fixed, thus minimizing losing of generalization. Although the advantages are remarkable, these methods often work with a very small learning rate, implicitly encouraging a small
deviation from the pre-trained CLIP, limiting the segmentation performance. In a nutshell, the trade-off between preserving CLIP's generalization and learning segmentation knowledge persists, hindering the final performance. Based on the paradigm of existing fine-tuning-based methods, our method explores a better trade-off from a fresh viewpoint: hyperspherical space.

\subsection{Large-scale Model Fine-tuning} Along with the improvement of large-scale foundation models~\cite{VLM_BLIP_2022_ICML,VLM_vilbert_NIPS_2019,VLM_ground_cvpr_2022,SAM_ICCV_2023,VLM_GLIPV2_2022_nips,VLM_VL_BERT_arxiv_2019,VLM_FLAVA_CVPR_2022,vlm_LLAMA_2023_arxiv,minigpt_arxiv_2023}, e.g., segment anything model~\cite{SAM_ICCV_2023}, numerous fine-tuning works~\cite{SAM_parser_AAAI_2024,SAM-COBOT_CVPR_2024,SAM-adapter_ICCVW_2023,COCOOP_CVPR_2022,COOP_IJCV_2022,clip-adapter_IJCV_2024,FT_llama_2023_arxiv,FT_PMC_arxiv_2023,FT_vit_2024_nips,FT_fs_iccv_2023} are proposed to adapt these models to various downstream scenarios. The core of these approaches lies in updating only limited parameters to capture the specific characteristics of different scenarios, while keeping most parameters fixed to maintain generalization. In contrast, fine-tuning CLIP for open-vocabulary semantic segmentation often meets a dilemma. On the one hand, limited parameters typically fall short in facilitating the transition from a classification model, i.e., CLIP, to a segmentation task. On the other hand, directly increasing the number of trainable parameters risks undermining CLIP’s ability to generalize to unseen classes, as experimented in CAT-Seg~\cite{catseg_cvpr_2024}. Most methods~\cite{fcclip_nips_2023, SED_cvpr_2024} solve this issue by simply freezing CLIP's text encoder and fine-tuning its image encoder, inevitably causing misalignment between the two modalities of CLIP. In this paper, we shed light on how to preserve generalization in a symmetric parameter-efficient fine-tuning manner and strive to explore an appropriate fine-tuning method for open-vocabulary semantic segmentation.

\section{Preliminaries}

\subsection{Hyperspherical Energy}

Existing fine-tuning methods implicitly assume that a smaller Euclidean distance between the fine-tuned model and the pre-trained model indicates better preservation of the pre-trained ability. However, Euclidean distance is insufficient to fully capture the degree of semantic preservation. Inspired by the Thomson problem \cite{Tomson_1904}, which seeks to determine the minimum electrostatic potential energy configuration of \( N \) mutually-repelling electrons on the surface of a unit sphere, we adopt the \textit{Hyperspherical Energy} to characterize the diversity of the model.

The hyperspherical energy function of a fully connected layer \( \bm{W} \) is defined as:
\begin{equation}
    \text{HE}(\bm{W}) := \sum_{i \neq j} \|\hat{\bm{w}}_i - \hat{\bm{w}}_j\|^{-1} \nonumber
\end{equation}
\noindent where \( \hat{\bm{w}}_i := \bm{w}_i / \|\bm{w}_i\| \) denotes the \( i \)-th normalized neuron. The power of the model representation can be characterized by the hyperspherical energy of its neurons. Higher energy implies greater redundancy, while lower energy indicates that the neurons of the model are more diverse. 

To ensure that the generalization is not destroyed during fine-tuning, we hypothesize that a good fine-tuning model should minimize the difference in hyperspherical energy compared to the pre-trained model:
\begin{align} \label{eq:oft_principle}
&\min_{\bm{W}}  \left\| \text{HE}(\bm{W}) - \text{HE}(\bm{W}^0) \right\| \nonumber \\
\Leftrightarrow & \min_{\bm{W}} \bigg\| \sum_{i \neq j} \|\hat{\bm{w}}_i - \hat{\bm{w}}_j\|^{-1}  - \sum_{i \neq j} \|\hat{\bm{w}}^0_i - \hat{\bm{w}}^0_j\|^{-1} \bigg\|.
\end{align}
It is evident that the attainable minimum is zero for Eq. \eqref{eq:oft_principle}. In this case, the hyperspherical energy satisfies an invariance property: applying the same orthogonal transformation to all neurons preserves pairwise hyperspherical similarity. Consequently, the minimum of zero can be achieved as long as \( \bm{W} \) and \( \bm{W}^0 \) differ only by a rotation or reflection, i.e., 
$\bm{W} = \bm{R}\bm{W}^0$
where \( \bm{R} \in \mathbb{R}^{d \times d} \) is an orthogonal matrix (with determinant \( 1 \) for rotation or \( -1 \) for reflection).

\subsection{Notation of Tensor Product}
In this section, we introduce the fundamental concept to achieve dual cross relation communication (Sec.~\ref{dcrc}): tensor product. A p-order tensor is indexed by $p$ indices and can be represented as a multidimensional array of data. Formally, a p-order tensor $\mathcal{A}$ can be written as $\mathcal{A} = (a_{i_1,i_2,\cdots,i_p}) \in \mathbb{R}^{n_1 \times n_2 \times \cdots n_p}$. Slices of a tensor are matrices defined from the tensor by holding all but two indices constant. For a $3$-order tensor, $\mathcal{A}(:,:,k)$ corresponds the $k^{\text{th}}$ frontal slice. For $p$-order tensors, matrix slices of $p$-order tensors can be referenced using linear indexing by reshaping the tensor into an $n_1 \times n_2 \times n_3 n_4 \cdots n_p$ $3$-order tensor and referring to the $k^\text{th}$ frontal slice as $\mathcal{A}(:,:,k)$. $\mathcal{A}_i \in \mathbb{R}^{n_1 \times n_2 \times \cdots n_{p-1}}$ for $i = 1, \cdots, n_p$ denotes the $(p-1)$-order tensor created by holding the $p$th index of $\mathcal{A}$ fixed at $i$. It is possible to create a tensor in a block circulant pattern, where each block is a tensor of $(p-1)$-order:
\begin{equation}
    \text{circ}(\mathcal{A}) = \begin{bmatrix}
        \mathcal{A}_{1}& \mathcal{A}_{n_p} & \mathcal{A}_{n_p-1} &\cdots & \mathcal{A}_{2}  \\
         \mathcal{A}_{2}& \mathcal{A}_{1} & \mathcal{A}_{n_p} &\cdots & \mathcal{A}_{3} \\
        \vdots           &     \vdots     & \vdots    & \ddots &   \vdots         \\
        \mathcal{A}_{n_p}& \mathcal{A}_{n_p-1} & \mathcal{A}_{n_p-2} &\cdots & \mathcal{A}_{1} 
    \end{bmatrix},
\end{equation}
where $\text{circ}(\cdot) $ creates a block circulant tensor and the size of $\text{circ}(\mathcal{A}) $ is $(n_1 n_p \times n_2 n_p \times \cdots \times n_{p-2} n_p \times n_{p-1})$. define $\text{unfold}(\cdot)$ to take an $n_1 \times \cdots \times n_p$ tensor $\mathcal{A}$ and return an $n_1 n_p \times n_2 \times \cdots n_{p-1}$ block tensor in the following way:
\begin{equation}
    \text{unfold}(\mathcal{A}) = \begin{bmatrix}
        \mathcal{A}_1 & \mathcal{A}_2 & \cdots & \mathcal{A}_{n_p}
    \end{bmatrix}^T. 
\end{equation}
The operation that takes $\text{unfold}$ back to tensor form is the ``$\text{fold}$'' command. Specially, $\text{fold}(\cdot, n_p)$ takes an $n_1 n_p \times n_2 \times \cdots \times n_{p-1}$ block tensor and returns an $n_1 \times \cdots \times n_p$ tensor, defined as:
\begin{equation}
    \text{fold}(\text{unfold}(\mathcal{A}), n_p) = \mathcal{A}.
\end{equation}

\begin{figure*}[t]
  \centering
  \small
  \includegraphics[width=\textwidth]{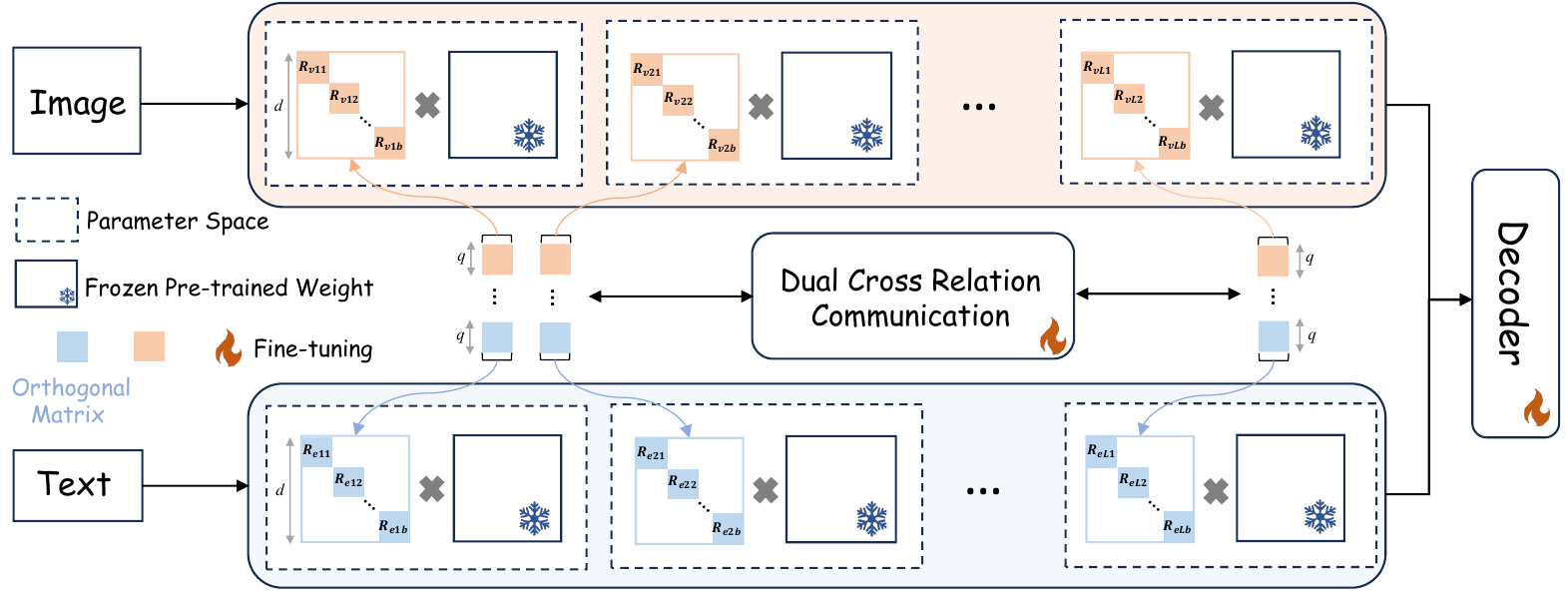}
  \caption{\textbf{A schematic representation of \alg.} In the \alg framework, we propose a partial orthogonal fine-tuning strategy, where each pre-trained weight matrix is paired with a tuned block-diagonal transformation matrix, some of which are orthogonal to preserve generalization. Then, we introduce a dual cross-relation communication mechanism to facilitate communication among all matrices, enabling alignment between different modalities.}
  \label{fig_big_picture}
\end{figure*}
\section{Methodology}

\subsection{Overview of \textbf{\alg}}
Fig.~\ref{fig_big_picture} illustrates the proposed \alg framework, which is based on two core components: (1) \blg updates the pre-trained parameter space of CLIP using a series of block-diagonal transformation matrices. According to analysis in Sec.~\ref{sec.1}, each parameter matrix in CLIP's text encoder is orthogonal to preserve generalization. (2) DCRC incorporates cross-modal and cross-layer communication within all tunable matrices, facilitating alignment between different modalities.

\subsection{Partial Orthogonal Parameterization}

The core idea of partial orthogonal parametrization (\blg) is to introduce the concept of hyperspherical space for fine-tuning CLIP. In this hyperspherical space, we fine-tune CLIP's text encoder under an orthogonality design principle from OFT~\cite{OFT_nips_2023} to preserve the hyperspherical energy of the pre-trained parameter space. Similarly, we use Cayley parameterization~\cite{cayley_1846} to ensure a tunable matrix $\bm{R}$ is strictly orthogonal, formally as:
\begin{equation}
\bm{R}=(\bm{I}+\bm{Q})(\bm{I}-\bm{Q})^{-1},    
\end{equation}
where $\bm{Q}$ is skew-symmetric. Guided by the observation in~\cite{Lseg_ICLR_2022, OVSeg_CVPR_2023} that CLIP's generalization ability is primarily preserved in its text encoder, we impose an orthogonality constraint on the tunable matrices within CLIP's text encoder during fine-tuning, which is defined as:
\begin{equation}
\bm{R}^\top\bm{R}=\bm{R}\bm{R}^\top=\bm{I},    
\end{equation}
where $\bm{I}$ is an identity matrix. Considering the relatively large dimension $d$ of the pre-trained matrix, for better efficiency, we introduce a block-diagonal structure by parameterizing $\bm{R}$ with $b$ blocks, formally as: 
\begin{align}
    \bm{R}&=\text{diag}(\bm{R}_1,\bm{R}_2,\cdots \bm{R}_i,\cdots,\bm{R}_b) \nonumber \\
    &=\begin{bmatrix}
    \bm{R}_1 &  &\\
    &\ddots & \\
    & & \bm{R}_b
    \end{bmatrix},
\end{align}
where $\bm{R}_i\in\mathbb{R}^{d / b\times d / b}$. Specifically, denote $\mathcal{R}^{V} = \{\bm{R}_{v1}, \cdots, \bm{R}_{v\ell}, \cdots, \bm{R}_{vL}\}$ and $\mathcal{R}^{E} = \{\bm{R}_{e1}, \cdots, \bm{R}_{e\ell}, \cdots, \bm{R}_{eL}\}$ as the sets of block-diagonal matrices in CLIP's image encoder and text encoder, respectively, where $L$ is its number of Transformer layers, $\bm{R}_{v\ell} \in \mathbb{R}^{d_v \times d_v}$, and $\bm{R}_{e\ell} \in \mathbb{R}^{d_e \times d_e}$. In practice, the dimensions of the tunable matrix $\bm{R}_{vi} \in \mathbb{R}^{d_v \times d_v}$ and $\bm{R}_{ei} \in \mathbb{R}^{d_e \times d_e}$ are often not equal, e.g., $d_v=768$ and $d_v=512$ for ViT/16 version of CLIP. Considering that the dimension of each individual slice in a higher-order tensor must be consistent according to the tensor theorem, we use a block diagonal structure to align the matrix dimensions between the two modalities, and thus, it cannot be discarded. For notation simplicity, we set $d_v = d_e = d$.

\subsection{Dual Cross Relation Communication}
\label{dcrc}
Although in \blg, we relax the orthogonal constraint for CLIP's image encoder to learn segmentation capability, each layer of the image encoder still incorporates a limited number of parameters, which largely restricts the flexibility of the projection adjustment due to the limitation of Hidden Markov Chain along layers~\cite{RIB_2021_nips,CVIB_2020_nips,SAM-COBOT_CVPR_2024}. 
To address this limitation, one might consider fully fine-tuning instead of using a small number of parameters. However, this approach can cause a misalignment between image and text features in CLIP, resulting in sub-optimal performance~\cite{fcclip_nips_2023}. Based on the above analysis, we introduce Dual Cross-Relation Communication (DCRC), which facilitates interaction among different layers and modalities (i.e., text and image). DCRC explicitly enhances the flexibility of fine-tuned projection adjustments and prevents misalignment issues.

DCRC introduces cross-layer and cross-modality communication among different block-diagonal matrices, achieved through two relation projections. To do this, we first treat all blocks in ${\ell}^{\text{th}}$ layer as an individual slice in this $3$-order tensor $\mathcal{T}_\ell$, which is derived as follows:
\begin{align}
        \mathcal{T}_\ell = [\bm{R}_{v\ell1}, \bm{R}_{e\ell1}, &\cdots, \bm{R}_{v\ell i}, \bm{R}_{e\ell i}, \nonumber \\
        &\cdots,\bm{R}_{v\ell b}, \bm{R}_{e\ell b}] \in \mathbb{R}^{q \times q \times (b + b)}, \label{eq.4.3-1}
\end{align}
where $q = d / b$. Then, we treat the tensor $\mathcal{T}_\ell$  as an individual slice within a $4$-order tensor $\mathcal{T}$, defined as follows:
\begin{equation}
    \mathcal{T} = [\mathcal{T}_1,\mathcal{T}_2,\cdots,\mathcal{T}_\ell,\cdots,\mathcal{T}_L] \in \mathbb{R}^{q \times q \times (b + b) \times L}. \label{eq.4.3-2}
\end{equation}
Initially, according to the characteristics of gradient propagation in deep learning theory, i.e., chain rule, each frontal slice $\bm{R}_{\cdot \ell i} \in \{\mathbb{R}^{q \times q}\}^{(b+b) \times L}$ is updated sequentially in CLIP's encoder. As a result, updating the $\mathcal{T}$ lacks cross-frontal-slice communication, limiting the flexibility of adjusting fine-tuned projection. To address this, we introduce two special tensor products, i.e., \textbf{$3$-order T-product} and \textbf{Higher-order T-product}.

\textbf{Definition 4.1(3-order T-product)} For $\mathcal{A} \in \mathbb{R}^{n_1 \times n_2 \times n_3}$ and $\mathcal{B} \in \mathbb{R}^{n_2 \times l \times n_3 }$, the $3$-order T-product $\mathcal{C} \in \mathbb{R}^{n_1 \times l \times \times n_3 } = \mathcal{A} * \mathcal{B}$ is defined as:
\begin{equation}
    \mathcal{C} = \mathcal{A} * \mathcal{B} = \text{fold}(\text{circ}(\mathcal{A}) \cdot \text{unfold}(\mathcal{B})), \label{eq.4.3-3}
\end{equation}
where `` * '' represents a tensor product, ``$\cdot$'' represents standard matrix product.

\textbf{Definition 4.2(Higher-order T-product)} For $\mathcal{A} \in \mathbb{R}^{n_1 \times n_2 \times n_3 \cdots \times n_p}$ and $\mathcal{B} \in \mathbb{R}^{n_2 \times l \times n_3 \times \cdots \times n_p}$, the High-order T-product $\mathcal{C} \in \mathbb{R}^{n_1 \times l \times n_3 \cdots \times n_p} = \mathcal{A} * \mathcal{B}$ is defined as:
\begin{equation}
    \mathcal{C} = \mathcal{A} * \mathcal{B} = \text{fold}(\text{circ}(\mathcal{A}) * \text{unfold}(\mathcal{B})). \label{eq.4.3-4}
\end{equation}
If $\mathcal{A} \in \mathbb{R}^{n_1 \times n_2 \times n_3}$, according to the \textbf{3-order T-product}, there is an invertible transform $S_3(\cdot): \mathbb{R}^{n_1 \times n_2 \times n_3} \rightarrow \mathbb{R}^{n_1 \times n_2 \times n_3}$ in third dimension and it transform the Eq.~\eqref{eq.4.3-3} as:
\begin{align}
    \mathcal{C} &= S_3^{-1}(S_3(\mathcal{A}) \odot S_3(\mathcal{B})) \nonumber \\
    &= S_3^{-1}(\bar{\mathcal{A}} \odot \bar{\mathcal{B}}) = S_3^{-1}(\bar{\mathcal{C}}), \label{eq.4.3-5}
\end{align}
where $\bar{\mathcal{C}} = \bar{\mathcal{A}} \odot \bar{\mathcal{B}}$ denotes the frontal-slice-wise product (Definition 2.1 refers to~\cite{TTP_2015_LAA}) $\bar{\mathcal{C}}(;,;,i) = \bar{\mathcal{A}}(;,;,i) \cdot \bar{\mathcal{B}}(;,;,i), i = 1,2,\cdots, n_3$ and $S_3^{-1}(\cdot)$ is the inverse transform of $S_3(\cdot)$. According to the definition of the frontal-slice-wise product, the invertible transform $S_3(\cdot)$ is formulated as:
\begin{equation}
    \bar{\mathcal{A}} = S_3(\mathcal{A}) = \mathcal{A} \times_3 \mathbf{S}_3, \label{eq.4.3-6}
\end{equation}
where ``$\times_3$'' denotes the mode-3 product and $\mathbf{S}_3 \in \mathbb{R}^{n_3 \times n_3}$ is an arbitrary invertible matrix. Similarly, the inverse transform of Eq.~\eqref{eq.4.3-6} is derived as:
\begin{equation}
    \mathcal{A} = S^{-1}_3(\bar{\mathbf{A}}) = \bar{\mathcal{A}} \times_3 \mathbf{S}_3 
            ^{-1}. \label{eq.4.3-7}
\end{equation}
Similarly, if $\mathcal{A} \in \mathbb{R}^{n_1 \times n_2 \times \cdots \times n_p}$, according to the \textbf{Higer-order T-product}, there are invertible transform $S_i(\cdot): \mathbb{R}^{n_1 \times n_2 \times \cdots \times n_p} \rightarrow \mathbb{R}^{n_1 \times n_2 \times \cdots \times n_p}, i = 3,4,\cdots,p$ in $i^\text{th}$ dimension and they transform the Eq.~\eqref{eq.4.3-4} as:
\begin{equation}
    \mathcal{C} = \tilde{S}^{-1}(\tilde{S}(\mathcal{A}) \odot \tilde{S}(\mathcal{B})) = \tilde{S}^{-1}(\bar{\mathcal{A}} \odot \bar{\mathcal{B}}) = \tilde{S}^{-1}(\bar{\mathcal{C}}), \label{eq.4.3-8}
\end{equation}
where $\tilde{S}(\mathcal{A}) = S_p(S_{p-1}(\cdots S_3(\mathcal{A}) \cdots))$, $\bar{\mathcal{C}} = \bar{\mathcal{A}} \odot \bar{\mathcal{B}}$ denotes the frontal-slice-wise product $\bar{\mathcal{C}}(;,;,i) = \bar{\mathcal{A}}(;,;,i) \cdot \bar{\mathcal{B}}(;,;,i), i = 1,2,\cdots, n_3 n_4 \cdots n_p$ and $\tilde{S}^{-1}(\cdot)$ is the inverse transform of $\tilde{S}(\cdot)$. Similarly, the inverse transform $\tilde{S}(\cdot)$ is formulated as:
\begin{equation}
    \bar{\mathcal{A}} = \tilde{S}(\mathcal{A}) = \mathcal{A} \times_3 \mathbf{S}_3 \times_4 \mathbf{S}_4 \cdots \times_p \mathbf{S}_p, \label{eq.4.3-9}
\end{equation}
and its inverse transform is derived as:
\begin{equation}
    \mathcal{A} = \tilde{S}^{-1}(\bar{\mathcal{A}}) = \bar{\mathcal{A}} \times_3 \mathbf{S}_3^{-1} \times_4 \mathbf{S}_4^{-1} \cdots \times_p \mathbf{S}_p^{-1}. \label{eq.4.3-10}
\end{equation}

\noindent \textit{Derivation.} please refer to supplementary material.
\hfill $\blacksquare$

According to Eqs.~\eqref{eq.4.3-8},~\eqref{eq.4.3-9} and~\eqref{eq.4.3-10}, we adopt its idea and design arbitrary invertible relation matrix $\mathbf{S}_3 \in \mathbb{R}^{(b + b) \times (b + b)}$ and $\mathbf{S}_4 \in \mathbb{R}^{L \times L}$ to capture the cross-modality and cross-layer information in $\mathcal{T}$. Then the updated tensor $\mathcal{T}_w$ is formulated as:
\begin{equation}
    \mathcal{T}_w = \mathcal{T} \times_3 \mathbf{S}_3 \times_4 \mathbf{S}_4 \in \mathbb{R}^{q \times q \times (b + b) \times L}, \label{eq.4.3-11}
\end{equation}
where the relation matrix $\mathbf{S}_3$ and $\mathbf{S}_4$ are learnable. To better capture the nonlinear interactions inside the whole parameter space, we further adopt $k$ layers deep neural network (DNN) $f_3(\cdot)$ and $f_4(\cdot)$ to replace the transform $\times_3 \mathbf{S}_3$ and $\times_4 \mathbf{S}_4$, respectively, and the DNN $f_3(\cdot)$ is formulated as:
\begin{equation}
    f_3(\mathcal{T}) = \sigma(\cdots  \sigma(\mathcal{A}\times_3\mathbf{W}_1)\cdots)\times_3  \mathbf{W}_k, \label{eq.4.3-12}
\end{equation}
where $\sigma(\cdot)$ is a nonlinear scalar function and matrices $\{ \mathbf{W}_j \in \mathbb{R}^{(b + b)} \}_{j=1}^k$. The DNN $f_4(\cdot)$ is similar. Finally, the $\mathcal{T}$ is updated by $\mathcal{T} = \mathcal{T} + \bm{\alpha}\mathcal{T}_w$, where $\bm{\alpha} \in \mathbb{R}^{(b+b) \times L}$ is a learnable parameter.

\subsection{Overall Architecture}

Overall, we develop a \alg framework, and for an input feature map $\textbf{M}_{\ell}$ in the ${\ell}^{\text{th}}$ Transformer layer of CLIP, the adjusted feature map provided by \alg, $\tilde{\textbf{M}}_{\ell}$, is formally via:
\begin{equation}\label{eq:coft_general}
\tilde{\textbf{M}}_{\ell} =\mathcal{F}_{\ell}(\textbf{M}_{\ell}; \mathcal{T}_\ell \mathbf{W}_{\ell}).
\end{equation}
where $\mathbf{W}_{\ell}$ is a pre-trained weight matrix in ${\ell}^{\text{th}}$ layer of CLIP's encoder, $\mathcal{T}_{\ell}$ is a 3-order tensor that is comprised of all the matrices $\boldsymbol{R}$ in $\ell^{th}$ layer, and $\mathcal{F}_{\ell}$ represents ${\ell}^{\text{th}}$ layer of CLIP's encoder. 

\noindent\textbf{Fine-tuning Stage.} During the fine-tuning phase, \alg is fine-tuned in conjunction with the original parameter space of CLIP, which is loaded from the pre-trained checkpoint and remains frozen.

\noindent\textbf{Loss Function.} Following previous works~\cite{catseg_cvpr_2024,SED_cvpr_2024}, we incorporate a cross-entropy loss, denoted as $\mathcal{L}_{\text{ce}}$, for the fine-tuning of CLIP.

\section{Experiments}
\label{experiments}
\begin{table*}[!t]
    \begin{center}
        \small
        \tabcolsep=0.08cm   
        \begin{tabular}{l|ccc|ccccc|c}
            \toprule

            Model & VLM & Additional Backbone & Fine-tuning Space & \texttt{A-847} & \texttt{PC-459} & \texttt{A-150} & \texttt{PC-59} & \texttt{PAS-20} & \texttt{PAS-20$^b$} \\
            \midrule
            \multicolumn{10}{c}{\textbf{\CC{15} \emph{Traditional Fine-Tuning}}} \\
            ZS3Net~\citep{z3c}  & - & ResNet-101  & E & - & - & - & 19.4 & 38.3 & - \\
            LSeg~\citep{Lseg_ICLR_2022} & CLIP ViT-B/32 & ResNet-101   & E & - & - & - & - & 47.4 & - \\
            ZegFormer~\citep{ZegFormer_CVPR_2022} & CLIP ViT-B/16 & ResNet-101   & E & 4.9 & 9.1 & 16.9 & 42.8 & 86.2 & 62.7 \\
            ZSseg~\citep{ZSseg_ECCV_2022} & CLIP ViT-B/16 & ResNet-101  & E & 7.0 & - & 20.5 & 47.7 & 88.4 & - \\
            OpenSeg~\citep{Openseg_ECCV_2022} & ALIGN & ResNet-101  & E & 4.4 & 7.9 & 17.5 & 40.1 & - & 63.8 \\
            OVSeg~\citep{OVSeg_CVPR_2023} & CLIP ViT-B/16 & ResNet-101c  & E & 7.1 & 11.0 & 24.8 & 53.3 & 92.6 & - \\
            ZegCLIP~\citep{segclip_CVPR_2023} & CLIP ViT-B/16 & -  & E & - & - & - & 41.2 & 93.6 & - \\
            SED~\cite{SED_cvpr_2024} & CLIP ConvNeXt-B & - & E &11.4 & 18.6 & 31.6 & 57.3 & 94.4 & - \\

            CAT-Seg~\cite{catseg_cvpr_2024} & CLIP ViT-B/16 & - & E &\underline{12.0} & \underline{19.0} & \underline{31.8} & \underline{57.5} & \underline{94.6} & \underline{77.3} \\ 

            \multicolumn{10}{c}{\textbf{\CC{15} \emph{Parameter-efficient Fine-Tuning}}} \\
            SAN~\citep{SAN_CVPR_2023} & CLIP ViT-B/16 & Side Adapter   & E & 10.1 & 12.6 & 27.5 & 53.8 & 94.0 & - \\

            Ours & CLIP ViT-B/16 & - & \textbf{H} & \textbf{12.5} & \textbf{19.4} & \textbf{32.4} & \textbf{57.9} & \textbf{95.2} & \textbf{78.2} \\
            \midrule
            \multicolumn{10}{c}{\textbf{\CC{15} \emph{Traditional Fine-Tuning}}} \\
            LSeg~\citep{Lseg_ICLR_2022} & CLIP ViT-B/32 & ViT-L/16 & E & - & - & - & - & 52.3 & - \\
            OpenSeg~\citep{Openseg_ECCV_2022} & ALIGN & Eff-B7  & E & 8.1 & 11.5 & 26.4 & 44.8 & - & 70.2 \\
            OVSeg~\citep{OVSeg_CVPR_2023} & CLIP ViT-L/14 & Swin-B & E & 9.0 & 12.4 & 29.6 & 55.7 & 94.5 & - \\
            SAN~\citep{SAN_CVPR_2023} & CLIP ViT-L/14 & -   & E & 12.4 & 15.7 & 32.1 & 57.7 & 94.6 & - \\
            ODISE~\citep{ODISE_CVPR_2023} & CLIP ViT-L/14 & Stable Diffusion  & E & 11.1 & 14.5 & 29.9 & 57.3 & - & - \\
            SED~\cite{SED_cvpr_2024} & CLIP ConvNeXt-L & - & E &13.9 & 22.6 & 35.2 & 60.6 & 96.1 & \underline{-} \\ 
            FC-CLIP~\cite{fcclip_nips_2023} & CLIP ConvNeXt-L & - & E &14.8 & 18.2 & 34.1 & 58.4 & 95.4 & - \\ 
            CAT-Seg~\cite{catseg_cvpr_2024} & CLIP ViT-L/14 & - & E & \underline{16.0} & \underline{23.8} & \underline{37.9} & \underline{63.3} & \underline{97.0} & \underline{82.5} \\

            \multicolumn{10}{c}{\textbf{\CC{15} \emph{Parameter-efficient Fine-Tuning}}} \\
            SAN~\citep{SAN_CVPR_2023} & CLIP ViT-L/14 & Side Adapter   & E & 12.4 & 15.7 & 32.1 & 57.7 & 94.6 & - \\
            Ours & CLIP ViT-L/14 & - & \textbf{H} & \textbf{16.5} & \textbf{24.2} & \textbf{38.4} & \textbf{64.1} & \textbf{97.7} & \textbf{83.2} \\

            \bottomrule
        \end{tabular}
        \caption{\textbf{Comparison with state-of-the-art methods on standard benchmarks.} The best-performing results are presented in bold, while the second-best results are underlined. ``E'': Euclidean Space. ``H'': Hypersphere Space.}
        \label{tab:main_table}
    \end{center}
\end{table*}

\subsection{Experimental Setup}

\noindent\textbf{Datasets.}
Following previous studies~\cite{catseg_cvpr_2024, SED_cvpr_2024}, we utilizes the COCO-Stuff dataset~\cite{COCO_2018_CVPR} as our training set. This dataset comprises approximately 118,000 densely annotated images across 171 distinct semantic categories. During inference, we carry out comparisons with state-of-the-art methods across several semantic segmentation datasets, including ADE20K \cite{ADE20K_IJCV_2019}, PASCAL VOC \cite{PASCAL_VOC_IJCV_2010}, and PASCAL-Context \cite{PASCAL_text_CVPR_2014}. 
\begin{itemize}
 \item\textbf{ADE20K~\cite{ADE20K_IJCV_2019}} is a classical semantic segmentation dataset comprising around 20,000 training images and 2,000 validation images. Besides, it includes two different test sets: \texttt{A-150} and \texttt{A-847}. The test set \texttt{A-150} has 150 common categories, while the test set \texttt{A-847} has 847 categories. 

 \item\textbf{PASCAL VOC~\cite{PASCAL_VOC_IJCV_2010}} is a small dataset for semantic segmentation, which includes 1464 training images and 1449 validation images. The dataset contains 20 different foreground categories. We name it as \texttt{PAS-20}. In line with~\cite{catseg_cvpr_2024}, we also report a score on \texttt{PAS-20$^b$}, which involves ``background'' as the 21st category. 

 \item\textbf{PASCAL-Context~\cite{PASCAL_text_CVPR_2014}} is upgraded from the original PASCAL VOC dataset. It includes two different test sets: \texttt{PC-59} and \texttt{PC-459} for evaluation. The test set \texttt{PC-59} has 59  categories, while the test set \texttt{PC-459} has 459 categories.
\end{itemize}

\noindent\textbf{Evaluation metric.} Following prior works~\cite{catseg_cvpr_2024,SED_cvpr_2024}, we adopt mean Intersection over Union (mIoU) to evaluate the semantic segmentation performance on the three benchmarks.

\noindent\textbf{Implementation Details.} We implement our method using the Transformer-based CLIP model. Following the protocol established in~\cite{catseg_cvpr_2024}, we evaluate our results on two versions of the CLIP model: ViT-B/16 and ViT-L/14. For training, we use the Adam optimizer~\cite{adam_arxiv_2014} with an initial learning rate of $5 \times 10^{-6}$ for CLIP, and a weight decay of $10^{-4}$. Training is conducted with one image per mini-batch. We set $q = 128$ for balancing efficiency and performance. The function $f_3(\cdot)$ and $f_4(\cdot)$ are implemented using two 2-layer MLPs. We act the cost-based approach provided in~\cite{catseg_cvpr_2024} as our decoder. All models are trained over 80,000 iterations on 4 NVIDIA RTX 3090 GPUs.

\subsection{Main Results}

\begin{table}[h]
    \small
    \centering
    \setlength{\tabcolsep}{3pt}
    \renewcommand{\arraystretch}{1.22}
    \begin{tabular}{l|cccc}
        \toprule
        Methods & OVSeg~\cite{OVSeg_CVPR_2023} & CAT-Seg~\cite{catseg_cvpr_2024} & SAN~\cite{SAN_CVPR_2023} & Ours \\ 
        \midrule
        Param.~(M) & 147.2 & 25.6 & 8.4 & \textbf{5.6} \\ 
        \bottomrule
    \end{tabular}
    \caption{\small \textbf{Efficiency comparison} in terms of learnable parameters.} 
    \label{table:Efficiency comparison}
\end{table}

\textbf{Comparing to SOTAs.} Here, we compare our proposed \alg with several state-of-the-art methods, as shown in Table~\ref{tab:main_table}, using six test sets across three benchmarks. Overall, we achieve the best results. Most existing open-vocabulary semantic segmentation methods employ traditional fine-tuning approaches, i.e., full or partial fine-tuning (tuning certain layers of CLIP). While these methods offer sufficient flexibility for learning new knowledge, they often result in a significant performance drop on unseen classes, as observed with
OVSeg~\cite{OVSeg_CVPR_2023}. Among these methods, CAT-Seg~\cite{catseg_cvpr_2024} achieves performance comparable to ours. However, its fine-tuning scheme is manually controlled through different layer combinations, necessitating a careful design to balance generalization and flexibility, while ours does not suffer from such an issue. Then, compared to SAN~\cite{SAN_CVPR_2023}, another parameter-efficient fine-tuning method that introduces only a limited number of tunable parameters, our approach significantly outperforms it, achieving improvements of 6.6\% on the \texttt{PC-459} dataset and 3.9\% on the \texttt{PC-59} dataset with ViT-B/16 as the base model. These results demonstrate the effectiveness of our method in preserving generalization while learning segmentation knowledge.

\begin{figure*}[t]
  \centering
  \begin{overpic}[width=1.0\linewidth]{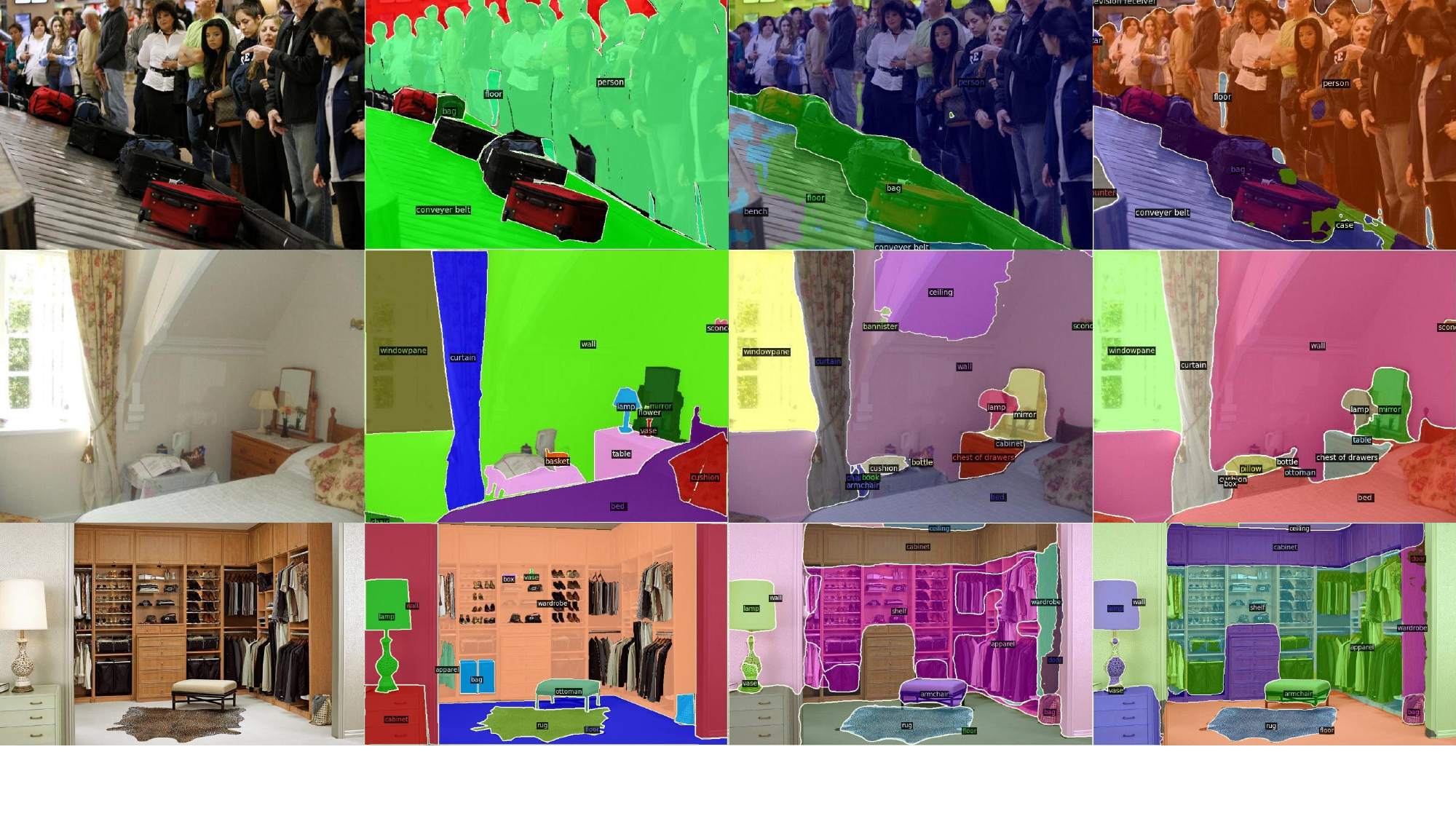}
   \put(10.0,2){\footnotesize{Image}}
   \put(30.0,2){\footnotesize{Ground truth}}
   \put(58.0,2){\footnotesize{CAT-Seg \cite{catseg_cvpr_2024}}}
   \put(85,2){\footnotesize{Ours}}
  \end{overpic}
  \caption{ Comparison of qualitative reults on ADE20K \cite{ADE20K_IJCV_2019} with 150 categories. We compare \alg with a state-of-the-art method, i.e., CAT-Seg \cite{catseg_cvpr_2024}.}
  \label{ade20k}
\end{figure*}

\noindent\textbf{Qualitative results.} Here, we visualize our method's representative example segmentation results against prevailing methods, e.g., CAT-Seg~\cite{catseg_cvpr_2024} in the \texttt{A-150} dataset. As shown in Figs.~\ref{ade20k}, we observe that our approach is able to generalize on diverse scenarios and produce more accurate results. More visualizations can be found in supplemental materials.

\begin{table*}[!t]
    \begin{center}
\small
\tabcolsep=0.1cm   
    \begin{tabular}{l|ccc|cccccc}
    \toprule

        Method & \blg & DCRC & Parameter counts~(M) & \texttt{A-847} & \texttt{PC-459} & \texttt{A-150} & \texttt{PC-59} & \texttt{PAS-20} & \texttt{PAS-20$^b$}\\
        \midrule\midrule

        Freeze & - & - & 0 & 4.4 & 6.6 & 24.8 & 49.4 & 92.5 & 71.9 \\ 
            \multicolumn{10}{c}{\textbf{\CC{15} \emph{Methods for Parameter-efficient Fine-Tuning}}} \\
    LoRA~\cite{PEFT_LORA_2022_ICLR}& - & - & 7.5 & 11.4 & 17.6 & 28.6 & 55.1 & 94.2 & 76.7 \\ OFT~\cite{OFT_nips_2023}& - & - & 3.8 & 10.9 & 18.0 & 30.2 & 53.7 & 93.7 & 74.3 \\ VPT~\cite{VPT_ECCV_2022}& - & - & 2.2 & 5.7 & 10.2 & 23.7 & 54.3 & 93.8 & 75.1 \\ Adapter~\cite{Adapter_ICML2019}& - & - & 7.5 & 10.4 & 16.5 & 28.8 & 54.9 & 94.2 & 75.2 \\ LST~\cite{LST_NIPS_2022}& - & - & 19.8 & 7.2 & 12.7 & 27.0 & 56.8 & 95.4 & 76.3 \\  SSF~\cite{SSF_NIPS_2022}& - & - & 0.6 & 6.9 & 15.2 & 28.6 & 52.1 & 93.2 & 72.8 \\ \midrule
            \multicolumn{10}{c}{\textbf{\CC{15} \emph{Ours}}} \\
         & \cmark & \xmark & 5.62 & 12.3 & 19.0 & 31.6 & 56.4 & 94.6 & 76.3 \\
        \alg & \xmark & \cmark & 0.01 & 7.6 & 10.9 & 26.8 & 53.6 & 92.7 & 74.5 \\
         & \cmark & \cmark & 5.63 & \textbf{12.5} & \textbf{19.4} & \textbf{32.4} & \textbf{57.9} & \textbf{95.2} & \textbf{78.2} \\

        \bottomrule
    \end{tabular}
        \caption{\textbf{Ablation study on the components of \alg} and universal PEFT methods, i.e.,``LoRA''~\cite{PEFT_LORA_2022_ICLR}, ``OFT''~\cite{OFT_nips_2023}, ``VPT''~\cite{VPT_ECCV_2022}, ``Adapter''~\cite{Adapter_ICML2019}, ``LST''~\cite{LST_NIPS_2022} and ``SSF''~\cite{SSF_NIPS_2022}. ``POP'': Partial Orthogonal Parameterization. ``DCRC'': Dual Cross Relation Communication. Towards easy to compare, we exclude the segmentation decoder when calculating the number of learnable parameters. The base model is ViT-B/16.}
    \label{tab:ablation_study}
    \end{center}

\end{table*}

\begin{table*}[!t]
    \begin{center}
\small
\tabcolsep=0.1cm   
    \begin{tabular}{c|cc|cccccc}
    \toprule

        & Block dimension $q$ & Parameter counts~(M) & \texttt{A-847} & \texttt{PC-459} & \texttt{A-150} & \texttt{PC-59} & \texttt{PAS-20} & \texttt{PAS-20$^b$}\\
        \cmidrule{2-9}
        & 256 $\times$ 256 & 22.52 & 12.4 & 19.2 & \textbf{32.7} & 57.6 & \textbf{95.4} & 77.9 \\
     (a)  & 128 $\times$ 128  & 5.63 & \textbf{12.5} & \textbf{19.4} & 32.4 & \textbf{57.9} & 95.2 & \textbf{78.2}  \\
       & 64 $\times$ 64  & 1.41 &  11.7 & 18.4 & 31.7 & 56.9 & 95.0 & 76.4 \\ \midrule

        & Orthogonal Constraint & Parameter counts~(M) & \texttt{A-847} & \texttt{PC-459} & \texttt{A-150} & \texttt{PC-59} & \texttt{PAS-20} & \texttt{PAS-20$^b$}\\
        \cmidrule{2-9}
        & w/o & 7.51 & 11.9 & 18.5 & 32.2 & 57.5 & \textbf{95.3} & 76.9 \\
     (b)  & with  & 3.76 & 12.2 & 19.1 & 31.4 & 57.1 & 94.3 & 76.8  \\
       & \blg & 5.63 & \textbf{12.5} & \textbf{19.4} & \textbf{32.4} & \textbf{57.9} & 95.2 & \textbf{78.2} \\

        \bottomrule
    \end{tabular}
        \caption{\textbf{Ablation study on different designs in \blg}. We show the impact of  (a) different block dimensions $q$ and (b) orthogonal constraints. The base model is ViT-B/16.}
    \label{tab:flexibility}
    \end{center}

\end{table*}

\noindent\textbf{Efficiency comparison.} We compare the efficiency of our method with other approaches, including OVSeg~\cite{OVSeg_CVPR_2023}, CAT-Seg~\cite{catseg_cvpr_2024}, and SAN~\cite{SAN_CVPR_2023}, all of which utilize CLIP ViT models. The comparison, summarized in Table~\ref{table:Efficiency comparison}, shows that our method employs the fewest trainable parameters while balancing the generalization of the pre-trained model and the flexibility for learning new knowledge. Additionally, since we introduce a lightweight architecture for calculating relations, specifically two relation matrices, the inference overhead is negligible during the inference phase.

\noindent\textbf{Generalization on other dense prediction tasks.} We validate our method on an open-vocabulary object detection task and experiment on COCO dataset following~\cite{CLIM_AAAI_2024} in Table~\ref{table:Comparison on OVD}. The results demonstrate that our method can generalize to mainstream dense prediction tasks, including segmentation and detection.

\begin{table}[h]
    \small
    \centering
    \setlength{\tabcolsep}{7pt}
    \renewcommand{\arraystretch}{1.22}
    \begin{tabular}{l|cc}
        \toprule
        Methods & \textbf{AP}$^{\text{Base}}_{50}$ & \textbf{AP}$^{\text{Novel}}_{50}$ \\  \midrule
        CLIP & 21.6 & 36.4  \\ 
        \multicolumn{3}{c}{\textbf{\CC{15} \emph{Traditional Fine-Tuning}}} \\
        CLIM~\cite{CLIM_AAAI_2024} & 25.7 & 42.5  \\
        \multicolumn{3}{c}{\textbf{\CC{15} \emph{Parameter-efficient Fine-Tuning}}} \\
        LoRA~\cite{PEFT_LORA_2022_ICLR} & 24.4 & 41.5  \\
        H-CLIP & 25.1 & 42.9  \\
        \bottomrule
    \end{tabular}
    \caption{\small \textbf{Comparisons on fine-tuning CLIP for an open-vocabulary object detection task} on the COCO dataset~\cite{COCO_dataset}. Results are measured by the box AP at IoU threshold 0.5.} 
    \label{table:Comparison on OVD}
\end{table}

\subsection{Ablative Studies}

\noindent\textbf{Ablation of Main Components.} Here, we conduct an ablation study to demonstrate the benefits of each component of our proposed \alg: partial orthogonal fine-tuning (\blg) and dual cross-relation communication (DCRC). We use the ViT-B/16~\cite{VIT_arxiv_2020} version of CLIP as the baseline, shown in row 1 of Table~\ref{tab:ablation_study}. Additionally, we implement a mainstream parameter-efficient fine-tuning (PEFT) method, LoRA~\cite{PEFT_LORA_2022_ICLR}, for comparison with a similar number of learnable parameters, as shown in row 2. Note that LoRA can improve performance compared to the baseline, demonstrating that PEFT is a viable approach for this task. However, the goal of traditional PEFT methods, which update weights using a small number of learnable parameters, does not necessarily preserve the generalization capability of pre-trained CLIP. In contrast, we introduce the principle of hyperspherical energy, which effectively ensures the minimum undermining of CLIP's intrinsic semantic structure, resulting in improved performance. Then, comparing row 5 to row 2, we observe significant performance gains, indicating that our results are driven by our targeted solution rather than merely the number of parameters. Moreover, row 3 shows that using only \blg preserves generalization on unseen classes, particularly in the \texttt{A-847} dataset. Meanwhile, solely adapting DCRC shows limited improvement, as it only enhances communication among frozen weight matrices. Finally, integrating DCRC with \blg yields clear performance gains, e.g., a 12.6\% improvement on the \texttt{PC-459} dataset.

\noindent\textbf{Discussion of \blg.}
Table~\ref{tab:flexibility} presents experiments introducing different designs into \blg. The design of \blg is related to (1) block dimension, i.e., $q$, and (2) how orthogonality constraints are applied. In (a), the results show that larger 
In (a), the results show that larger $q$ generally performs better than smaller $q$. However, we find a good trade-off between performance and parameter efficiency, with $q = 128$ working well across datasets and tasks. Therefore, we maintain this setting in other experiments. In (b), we show that both blindly applying orthogonality constraints to the learnable matrices of all layers and not using any constraints at all can degrade performance on several datasets, demonstrating the value of our analysis with the hyperspherical energy principle. Notably, when we adopts orthogonal constraints on both CLIP's image and text encoders to strictly maintain the original semantic structures, it is similar to equipping OFT~\cite{OFT_nips_2023} with our proposed DCRC module. Differently, our method applies orthogonal constraints only to the text encoder. This is crucial for open-vocabulary semantic segmentation, as it can provide more flexibility in fine-tuning the image encoder, facilitating the transfer of CLIP's initial alignment from image-level to pixel-level.

\section{Conclusion}

In this paper, we propose a \alg framework to address three issues: 1) high computational cost, 2) misalignment between the two inherent modalities of CLIP, and 3) degraded generalization ability on unseen categories when equipping CLIP with pixel-level prediction ability for open-vocabulary semantic segmentation. Specifically, we propose a symmetrical parameter-efficient fine-tuning (PEFT) strategy conducted in hyperspherical space for both of the two CLIP modalities. Specifically, the PEFT strategy is achieved by a series of efficient block-diagonal learnable transformation matrices and a dual cross-relation communication module among all learnable matrices to mitigate misalignment between different modalities. Furthermore, we apply an additional constraint to PEFT on the CLIP text encoder according to the hyperspherical energy principle, i.e., minimizing hyperspherical energy during fine-tuning preserves the intrinsic structure of the original parameter space, to prevent the destruction of the generalization ability offered by the CLIP text encoder. Extensive experiments demonstrate that the proposed \alg framework generalized improves segmentation performance across several benchmarks while introducing approximately 4\% of CLIP's total parameters. We hope our approach will provide a new direction and inspire future research in this field.


\clearpage
{
    \small
    \bibliographystyle{ieeenat_fullname}
    \bibliography{main}
}


\end{document}